\documentclass[10pt, conference, compsocconf]{IEEEtran}

\usepackage{paralist}
\usepackage{multirow}
\usepackage{amssymb}
\usepackage{color}
\usepackage{graphicx}

\usepackage{booktabs}
\usepackage{amsmath}
\usepackage{algorithm}
\usepackage[noend]{algpseudocode}
\usepackage{rotating}
\usepackage{bbm}
\usepackage{listings}

\usepackage{xcolor,colortbl}
\definecolor{Gray}{gray}{0.85}
\definecolor{aliceblue}{rgb}{0.94, 0.97, 1.0}
\newcolumntype{a}{>{\columncolor{Gray}}l}
\newcolumntype{b}{>{\columncolor{aliceblue}}l}

\newsavebox\CBox

\let\originalleft\left
\let\originalright\right
\renewcommand{\left}{\mathopen{}\mathclose\bgroup\originalleft}
\renewcommand{\right}{\aftergroup\egroup\originalright}

\makeatletter
\def\ps@IEEEtitlepagestyle{
  \def\@oddfoot{\mycopyrightnotice}
  \def\@evenfoot{}
}
\def\mycopyrightnotice{
  {\footnotesize
  \begin{minipage}{\textwidth}
  \centering
  \copyright~2017 IEEE. Personal use of this material is permitted. Permission from IEEE must be obtained for all other uses, in any current or future media, including reprinting/republishing this material for advertising or promotional purposes, creating new collective works, for resale or redistribution to servers or lists, or reuse of any copyrighted component of this work in other works.
  \end{minipage}
  }
}

\begin{document}
%
\title{Compact Multi-Class Boosted Trees}


\author{
	Natalia Ponomareva,
	Thomas Colthurst,
	Gilbert Hendry,
	Salem Haykal,
	Soroush Radpour \\
	Google, Inc. \\
	{\tt tfbt-public@google.com}
	}


%


\maketitle

\begin{abstract}
Gradient boosted decision trees are a popular machine learning technique,
in part because of their ability to give good accuracy with small models.
We describe two extensions to the standard tree boosting algorithm designed
to increase this advantage. The first improvement extends the boosting
formalism from scalar-valued trees to vector-valued trees. This allows
individual trees to be used as multiclass classifiers, rather than requiring
one tree per class, and drastically reduces the model size required for multiclass
problems. We also show that some other popular vector-valued gradient boosted
trees modifications fit into this formulation and can be easily obtained in our implementation. The second extension, layer-by-layer boosting,
takes smaller steps in function space, which is empirically shown to lead to a faster convergence and to a
more compact ensemble. We have added both improvements to the open-source TensorFlow Boosted
trees (TFBT) package, and we demonstrate their efficacy on a variety of multiclass datasets. We expect these extensions will be of particular interest
to boosted tree applications that require small models, such as embedded devices,
applications requiring fast inference, or applications desiring more interpretable
models.
\end{abstract}


%
\IEEEpeerreviewmaketitle

\section{Introduction}

There are many reasons to use and study gradient boosted decision trees (or
boosted trees for short).  They have outstanding accuracy, as demonstrated
by their winning performance
in numerous surveys of machine learning models (such as \cite{Caruana05anempirical})
and Kaggle competitions \cite{xgboost}.  They are easy to use, as input
features do not need to be whitened or otherwise normalized.  They are
flexible: by supporting custom loss functions, they can attack arbitrary
classification and regression problems, including ranking or regression
tasks.  They are backed by solid theory, which allows them to be viewed
as doing gradient descent in the space of functions by taking steps in the
form of decision trees \cite{mason2000boosting}.
And last but not least, they often produce compact models, with fewer parameters
for the same accuracy when compared to say random forests.  Those compact
models in turn lead to faster inference speeds, less memory consumption
(important for embedded devices and cellphones), and better interpretability.

This paper describes two extensions we've made to TensorFlow Boosted Trees
(TFBT) \cite{TfbtEcml} that are designed to increase the model compactness.
As its name suggests, TFBT is built on top of TensorFlow; it is open-source
and available on github under \texttt{tensorflow/contrib/boosted\_trees}.
The first extension, described in section \ref{multiclass},
extends the usual boosting theory to apply to vector-valued outputs.  Using
the derived update equations, we can attack multi-class classification
or multidimensional regression problems directly with trees storing
vectors in the leaves, rather than the 1-vs-rest or
1-vs-1 approaches that are commonly used \cite{Allwein:2001:RMB:944733.944737}.  Significantly
fewer trees are required for good performance with this approach, yielding
corresponding reductions in model complexity.

The second extension, layer-by-layer boosting, described in section \ref{layer-by-layer}, can be
thought of as taking smaller steps in function space when doing the
gradient descent; steps that correspond to tree layers (i.e., all the nodes
of equal depth) rather than entire trees.  These smaller steps let us
converge faster with fewer trees, especially with custom loss functions for
which a second-order Taylor expansion is inexact.

Finally, in section \ref{experiments}, we evaluate these two extensions.
On ensemble sizes up to 100 trees, we demonstrate that vector-valued trees lead to much faster convergence and smaller ensembles on multiclass datasets, and that the combination of layer-by-layer boosting and vector-valued
trees often produces significant performance improvements.

\section{Related work}
\label{label-related}
Many popular algorithms are inherently binary, for example, the extremely popular \textit{AdaBoost} \cite{Freund96experimentswith}. Historically if a multiclass dataset was used, this was addressed by decomposing the multiclass problem into a set of binary subproblems, namely either as 1-vs-rest (sometimes refered to as 1-vs-all, as well as tree-per-class) or 1-vs-1 (also referred to as ``all-pairs'' in some sources).

Even though binary reformulations achieve good performance on some datasets and are easy to implement, these approaches have a number of drawbacks. Firstly, the number of classifiers required is at least $C$ (the number of classes). Secondly, as Friedman et al. point out, even if decision boundaries between classes are simple, the decision boundaries to be learnt when the problem is reformulated into binary problems can become hard, thus making these boundaries difficult to approximate \cite{Friedman98additivelogistic}. Additionally, common 1-vs-rest can make the subproblems unbalanced, complicating the learning further \cite{lpboost}, since some losses like log-loss are susceptible to class imbalances. Furthermore, when applied to boosting, theoretical guarantees state that to achieve good performance, each of the weak learners must achieve an accuracy of at least 0.5  \cite{Zhu09multi-classadaboost}. This is comparable to random guess in case of binary classification with balanced classes, but for a multiclass problem with balanced classes, random guess would result only in $1/C$ accuracy \cite{Zhu09multi-classadaboost}.

Several attempts to tackle multiclass problems directly, by optimizing a loss that tries to classify all of the classes correctly at the same time, were made. One of the first implementation was \textit{LogitBoost} \cite{Friedman98additivelogistic} - a generalization of \textit{AdaBoost}. Friedman et al. showed that it produces results superior to those achieved by 1-vs-rest models on a simulated example with complicated intra-classes boundaries. \textit{LogitBoost} takes the multinomial logistic regression (cross-entropy loss) and decomposes it in a standard way of running $C$ independent binary logistic regressions, in which one label $m$ is chosen as ``pivot'' or base class, and the other $C-1$ labels are separately regressed against this pivot label. So for each class $j$, given a base class $m$, one can calculate a quasi-Newton update to improve the loss of class $j$ vs class $m$. Now Friedman et. al note that the choice of pivot $m$ is arbitrary, so they build a tree for $j$ by approximating an average of steps over all choices of base classes. \textit{LogitBoost} still works however by fitting $C$ trees during each boosting iteration, where each tree is derived to minimize the overall loss with regard to the $j$-th class - so essentially an ensemble of trees is built to predict scores for each class. An alternative view would be that Friedman et al. approximates the full Hessian matrix of the expanded (up to a second derivative) loss via diagonal approximation \cite{ICML2012Sun_569}. Also note that at each boosting step, the predictions are fixed - a new tree built during an iteration $i$ does not affect the other trees built during the same boosting iteration.

\textit{MART} \cite{Friedman00greedyfunction} is another popular variant of boosting with trees, which differs from \textit{LogitBoost} in that it uses only the first order gradient for finding the splits. Both \textit{MART} and \textit{LogitBoost} use second-order information to derive the weights on terminal nodes \cite{Sun:2014:CRA:3044805.3045032}. For multiclass, \textit{MART} adopts the same approach as \textit{LogitBoost}, building $C$ trees during each boosting iteration. An in-depth comparison between \textit{MART} and \textit{LogitBoost} is available in \cite{Sun:2014:CRA:3044805.3045032}.

An extension to \textit{LogitBoost} approach of multiclass handling is proposed in  \cite{DBLP:journals/corr/abs-1203-3491}, \cite{DBLP:journals/corr/abs-0811-1250}. Ping Li names it \textit{ABC-Boost} (with two implementations \textit{ABC-MART} and \textit{ABC-LogitBoost}, depending on whether the second order gradient was used for finding tree splits or not). He notes that for multiclass classification, for each point, the sum of scores for all classes can be required to equal a constant (due to the fact that adding a constant to the scores of all classes does not change the overall winning class and class probabilities when common softmax is applied to the scores). This requirement on the sum of the scores does not necessarily hold true for the \textit{LogitBoost} algorithm. If the constraint is enforced, the scores for only $C-1$ classes need to be calculated and the derivatives can be redefined. Additionally, instead of averaging over a choice of all ``pivot'' classes during each iteration, \textit{ABC-boost} adaptively chooses the best base class (by first considering all base classes and then choosing the one that maximally reduces the training loss). Experiments show improvements over the baseline algorithms, however a large number of boosting iterations (for example, around 4,000 on the covertype dataset) is still required to achieve good performance \cite{DBLP:journals/corr/abs-0811-1250}. Additionally experiments demonstrate that in some cases, \textit{ABC-MART} requires even more training iterations than the  original \textit{MART} algorithm \cite{DBLP:journals/corr/abs-0811-1250}. Nevertheless, it was shown that \textit{ABC-MART} and \textit{ABC-LogitBoost} improve over their respective original algorithms \cite{DBLP:journals/corr/abs-0811-1250}, as well as that \textit{ABC-LogitBoost} outperforms  \textit{ABC-MART} on most datasets. An alternative view of \textit{ABC-LogitBoost} is presented in \cite{ICML2012Sun_569}: Peng et al. show that the difference between \textit{LogitBoost} and \textit{ABC-LogitBoost} is essentially in the Hessian matrix approximation. In \textit{ABC-LogitBoost}, the Hessian is approximated by first choosing a dimension (``base'' or ``pivot'' class) and then again approximating the remaining $(C-1)$ x $(C-1)$ matrix using diagonal approximation, where in \textit{LogitBoost} the full $C$ x $C$ Hessian is approximated via diagonal. The results suggest that the better approximation in \textit{ABC-LogitBoost} results in improved performance \cite{ICML2012Sun_569}. Note that \textit{ABC-Boost} and its modifications still build $C-1$ trees at every iteration.

\textit{AOSO-LogitBoost} is an adaptation of \textit{LogitBoost} to multiclass problems where only one vector-leaf tree is built during an iteration. It essentially approximates the Hessian in a block-diagonal fashion, at each iteration selecting only scores for 2 classes to be updated, leading to a one-vs-one classifier \cite{ICML2012Sun_569}. \textit{AOSO-LogitBoost} is similar to \textit{ABC-LogitBoost} in that it also works only on a $(C-1)$ x $(C-1)$ matrix, where the final dimension, the ``pivot'', is fixed. The pair of classes is chosen based on the magnitude of the derivatives (i.e. choose the class you do the worst on). Experiments show that \textit{AOSO-LogitBoost} outperforms \textit{ABC-LogitBoost} and, not surprisingly, requires a smaller number of trees to reach convergence.

\textit{SAMME} \cite{Zhu09multi-classadaboost} is another attempt at tackling multiclass with boosting: it is a modification of \textit{AdaBoost} exponential loss (as opposed to multinomial logistic loss in \textit{LogitBoost}). \textit{SAMME} works on vector encoded labels, building only one tree per iteration. The class encoding scheme for the labels is as follows: instead of one-hot vector encoding, the authors use an encoding where 1 is put in position of the real class, and -1/(C-1) is set for all other classes positions. With this label encoding scheme, a new multi-class exponential loss is introduced. Optimizing this loss with a constraint that scores should sum to 0, \textit{SAMME} derives a closed-form solution by using a Lagrange multiplier and using only first-order derivatives (vs Newton update in \textit{LogitBoost}) with respect to the prediction so far and a Lagrange multiplier. \textit{SAMME} also shows that in this framework, weak classifiers are required to do better than $C$-random guessing.
 
\textit{GD-MCBoost} is another attempt at extending AdaBoost to a multiclass version \cite{NIPS2011_4450}, that also uses gradient descent step but a different encoding scheme for the labels. \textit{GD-MCBoost} has solid theory behind its encoding scheme and was shown to produce larger-margin models than \textit{SAMME}.

\textit{GAMBLE} \cite{doi:10.1137/1.9781611972771.27} also extends \textit{AdaBoost} to multiclass classification. Authors use the same multiclass exponential loss and label encoding scheme as in \textit{SAMME} and also produces one vector-leaf tree per boosting iteration. The difference between \textit{GAMBLE} and \textit{SAMME} lies in derivations for loss optimization - \textit{GAMBLE} uses quasi-Newton step (with second order derivative). The original paper \cite{doi:10.1137/1.9781611972771.27} provides evidence that \textit{GAMBLE} performs better on a number of datasets than both 1-vs-rest and \textit{SAMME}, which might be due to the fact that second order derivatives are used. 

Many more variants of multiclass vector-form boosting exists. However, most successful variations differ only in the following choices:
\begin{compactitem}
\item Choice of loss (cross-entropy/multinomial logistic, multiclass exponential etc.)
\item Choice of loss optimization (to use or not to use second order information,
\item If second order gradient is used, the choice of approximation of the Hessian.
\item The choice of enforcing the constraint on scores (via a penalty term added to the Loss, or not enforced, or enforced via softmax over the scores etc.)
\item And finally, choice of label encoding scheme (one hot, 1 vs -1/(C-1), etc.) 
\end{compactitem}

Many papers show that models that handle a multiclass loss directly and build trees with vector-leaves result in better convergence rates and better performing models. However, the adoption of these methods is unfortunately lacking. Many popular libraries like XGBoost \cite{xgboost}, Scikit-learn \cite{scikit-learn}, R GBM \cite{gbm}, LightGBM \cite{LightGBM} support multiclass only as 1-vs-rest, whereas Spark MLLib \cite{Meng:2016:MML:2946645.2946679} (at the time of writing this paper) does not support multiclass at all. This might be due to the fact that each modification that tackles multiclass problems uses different losses, weight update scheme and re-derives gradients and Hessians (if used), complicating implementations.
\section{Multiclass handling}
\label{multiclass}

In the conventional formulation that is used by many libraries like XGBoost \cite{xgboost}, gradient boosted trees store only scalar values
in their leaves. In order to handle vector regression or multiclass classification
problems, multiple scalar-leaved trees must be used.  In this section, we
show how the usual derivation of the update formula for gradient boosted
trees effortlessly generalizes to handling vector values which can handle
vector regression and multiclass problems directly.  For clarity,
we follow the derivation given in \cite{xgboost}.

Assuming $m$ is the number of features, $n$ is the number of instances and $C$ is the number of classes, our model maps the input $x_i \in \mathcal{R}^m, i \in 1..n $ to the output $\hat{y_i} \in \mathcal{R}^C$ using the sum of $K$ trees:
$$\hat{y_i} = \varphi(x_i) = \sum_{k=1}^K f_k(x_i)$$

Assuming $T$ is the number of leaves in a tree $f(x)$, we represent each tree $f(x)$ as the combination of a structure
function $q(x): \mathcal{R}^m \rightarrow 1..T$, mapping an instance to the tree leaf where it ends up, and a set of leaf weights
$\{ w_j \in \mathcal{R}^C \, | \, j \in 1..T \}$ so that $f(x) = w_{q(x)}$.  We seek to
minimize the regularized objective function
$$L(\varphi) = \sum_{i=1}^n l(y_i, \hat{y_i}) + \sum_{k=1}^K \Omega(f_k)$$
which is the sum of the non-regularized loss $l$ and the ensemble regularizer $\Omega$.
In what follows, we will specifically consider an $\Omega$ regularizer of the form
$$\Omega(f) = \alpha T + \frac{1}{2} \lambda \sum_{j=1}^T || w_j ||^2$$
that penalizes both the number of the tree leaves and the L2 norms of its leaf weight vectors.

The minimization is done iteratively, and the training happens in an additive manner: at the start of each boosting iteration $K$ we have $K-1$ fixed trees built so far and we are looking add in the new tree $f_K$ as to minimize
\begin{equation}
L^{(K)} = \sum_{i=1}^{n} l\left( y_i, \hat{y_i}^{(K-1)} + f_K(x_i) \right) + \Omega(f_K)
\label{obj-at-K}
\end{equation}

For a given function l(x), the vector form Taylor expansion (up to second order derivative) can be written as:

\begin{equation}
\label{label-taylor}
l(x+\Delta{x}) = l(x)+(\Delta{x})^\intercal \nabla{g(x)} + \frac{1}{2} (\Delta{x}^\intercal) H(x) \Delta{x}
\end{equation}
where $\nabla{g(x)}$ - is the vector of gradients, $H_g(x)$ is the matrix of second order derivatives. If x is C-dimensional, then $\nabla{g(x)}$ will be of size C and  $H(x)$  will be of size C x C.

By taking $f_K$ as a formal vector-valued symbol and applying formula \ref{label-taylor} to formula \ref{obj-at-K}, we get

\begin{align}
\begin{split}
L^{(K)} \approx \sum_{i=1}^{n} (l(y_i, \hat{y_i}^{(K-1)}) + f_K(x_i)^\intercal g_i + \\
\frac{1}{2}f_K(x_i)^\intercal H_i f_K(x_i)) + \Omega(f_K)
\end{split}
\end{align}

where $g_i$ is a $C$ size vector of gradients (with respect to the predicted score for the first class, second class etc.), $H_i$  - Hessian matrix (with respect to pairs of predicted scores, of size $C$ x $C$.

Now fix a structure $q(x_i)$ on $f_K$, and consider a particular weight vector
$w_j$ of leaf $j$.
We can rewrite the objective so the summation happens on leaf index j. Dropping the loss on predictions so far, which is constant, we get
\begin{align}
\begin{split}
L^{(K)}  = \sum_{j=1}^{T}(w_j ^\intercal \tilde{g_j} + \frac{1}{2}(w_j ^\intercal \tilde{H_j} w_j + \lambda w_j ^\intercal w_j)) + \gamma T \\
Leaf\_set_j = \{i \, | \, q(x_i) = j\} \\
\tilde{g_j} = \sum_{i \in Leaf\_set_j} g_i \\
\tilde{H}_j = \sum_{i \in Leaf\_set_j} H_i
\end{split}
\end{align}
where $\tilde{g_j}$ is the vector representing the sum of gradients of instances that fall into the leaf j, $\tilde{H_i}$ is the matrix representing the sum of Hessians of instances that fall into that leaf j. This objective is a sum of independent objectives per leaf. For each given leaf, we have
\begin{equation}
{\bar L}^{(K)}(w_j) = w_j^\intercal \tilde{g_j} + \frac{1}{2} w_j ^\intercal (\tilde{H_j}+ \lambda I)w_j
\end{equation}
where $I$ is the identity matrix.

This approximation is a quadratic function of the vector $w_j$ and
has a global minimum at
\begin{equation}
\label{eq-weight}
w_j = - \left(\lambda I + \tilde{H_j} \right)^{-1} \tilde{g_j} 
\end{equation}
if the matrix $\lambda I + \tilde{H_j}$ is symmetric positive definite.
This is easy
to guarantee:  by Schwarz's theorem, the individual Hessian matrices
$H_i$ will be
symmetric if the second-order derivatives are continuous, and a convex
loss function will make the $H_i$'s and their sum $\tilde{H_j}$ positive
semi-definite.
Finally, if $\lambda > 0$, adding $\lambda I$ to the sum will make the final
matrix positive definite.

Under those conditions, adding the leaf $j$ will decrease the loss by
\begin{equation}
\label{eq-gain}
\mbox{Gain}_j = \frac{1}{2} \tilde{g_j}^t
  (\lambda I + \tilde{H}_j)^{-1}
  \tilde{g_j} 
\end{equation}
To evaluate the quality of the 
split, the contributions of both the new left and right leaves $L$ and $R$
must be
compared against the contribution of the removed parent $P$, along with any
penalty the regularizer might impose on the increased tree complexity:
\begin{equation}
\label{eq-split-gain}
\mbox{Gain}_{P \rightarrow L,R} =
\mbox{Gain}_L + \mbox{Gain}_R - \mbox{Gain}_P - \gamma
\end{equation}
Similarly to XGBoost, we build our trees greedily based on this gain, and
always pick the split with the highest gain.  We also offer the usual option of
only picking a split if its gain is greater than zero ("pre-pruning"), or
allowing the gain to be negative
(which can happen because of the regularization)
and then post-pruning afterwards.
From an implementation point of view, it is worth noting that
$\tilde{g}_P = \tilde{g}_L + \tilde{g}_R$ and
$\tilde{H}_P = \tilde{H}_L + \tilde{H}_R$.

\subsection{Matrix inversion}
Both accumulation of Hessian matrices and matrix inversion in Formulas (\ref{eq-gain} and \ref{eq-weight}) are potentially expensive.
We implement two strategies in TFBT:
\subsubsection{Full Hessian}
Note that if we want to calculate $x = A^{-1} v$, or alternatively $Ax=v$, where $A$ is some matrix, instead of explicitly calculating
$A^{-1}$ and multiplying by $v$, we can treat it as linear least squares system \cite{golub}.
In general, the usual methods to solve such a system are SVD decomposition, the QR decomposition and normal equations.  SVD decomposition is accurate but slow, normal equations tend to be the least accurate but the fastest, and the QR decomposition is in between \cite{Lee_numericallyefficient}. We go with QR decomposition with column pivoting of $A$ using the \textit{Eigen} library.

\subsubsection{Diagonal Hessian}
One potential simplification would be to assume that matrix $\tilde{H_j}$ is diagonal, which makes both Hessian accumulation for each split and matrix inverse $O(C)$ (vs. $O(C^3)$ for full Hessian inverse and $O(C^2)$ for Hessians accumulation). In our experiments we show that such simplification actually does not result in decreased performance. On the contrary, it serves as additional regularization and results in faster convergence and better results.

\section{Choice of losses and extensions}
For classification problems with $C > 2$ classes, our default model stores
length $C$ vectors in the leaves, turns those vectors into a vector of $C$
probabilities using the softmax function, and then evaluates those probabilities
using the cross-entropy error function.  This model does have the drawback
of using $C$ parameters to describe the $C - 1$ dimensional space of
$C$ probabilities summing to $1$, but it has two out-weighing benefits:  it
maintains the symmetry between the class weights, and it allows the
regularization function $\Omega$ to maintain its usual form. In particular,
the zero vector of weights, which minimizes the L2 penalty, maps to the
maximally uninformative vector of equal probabilities, which is desirable.

This default model and loss is the one used for the experiments in section
\ref{experiments}.  However, as previously mentioned, our formulation
and implementation is easily extensible and can be used to obtain many of
the multiclass approaches discussed in Section \ref{label-related}, in the
following ways:
\begin{compactitem}
\item A compact (because we build 1 tree per iteration for all of the classes) version of \textit{LogitBoost}-like model can be obtained when multinomial logistic/cross entropy loss is chosen, and labels are encoded as 1-hot, with a diagonal Hessian mode.
\item \textit{SAMME}-like model can be approximated by not using the Hessian at all, and modifying the loss to add Lagrange to enforce the constraints of scores summing to 0.
\item \textit{GAMBLE}-like model can be obtained by using multiclass exponential loss, vector-form label encoding scheme of 1 in position of the real class, and -1/(C-1), and no regularization. The label encoding can be done in a TensorFlow input function.
\item \textit{AOSO LogitBoost} can be simulated by using cross-entropy loss and adding new block-coordinate strategy for Hessian approximation. 
\item \textit{GD-MCBoost} can be obtained by using the label encoding scheme from \cite{NIPS2011_4450}, no second order gradient in loss optimization and exponential multiclass loss.
\end{compactitem}

One thing to keep in mind is that for a large number of classes, the full Hessian approximation might become prohibitively slow. Since our experiments show no significant difference between full and diagonal Hessian, we recommend that diagonal Hessian is used as default for a larger number of classes.

Additionally, we would like to point out that our formulation should work with multi-label problems as well, as long as an appropriate multi-label loss is provided. However for multi-label problems it might be beneficial to explore other than diagonal approximations of the Hessian, that can account for the connections between the labels and the sparsity of the Hessian, for example block diagonal Hessians. It is however not suitable for extreme-multilabel problems, where only few labels out of millions apply to an instance. This is due to the fact that a full dense vector of scores will be stored in leaves.

Finally, we would like to highlight that since TensorFlow does automatic differentiation, switching between losses and creating new customizable losses for multiclass setting should be very easy in our proposed framework. Any twice-differentiable loss should be easily pluggable into TFBT.

\section{Layer-by-layer boosting}
\label{layer-by-layer}
One of the novel features of TFBT's tree building procedure is layer-by-layer boosting. In TFBT's layer-by-layer boosting, we allow internal nodes to contribute weight updates while the current tree is getting built. A leaf node's final contribution is therefore the aggregate contributions from its ancestors all the way up to the root node.
This is illustrated in the following diagram:
\begin{figure}[h!]
  \caption{Layered Prediction.}
  \centering
  \includegraphics[scale=0.39]{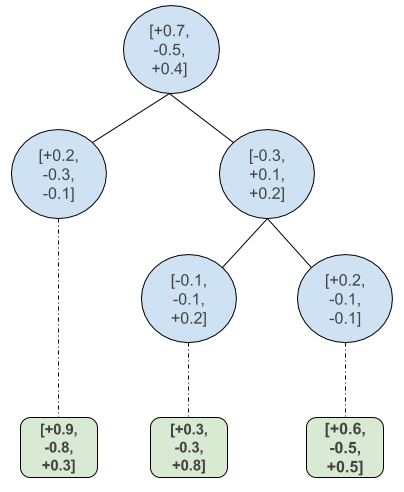}
  \label{figure:layered-prediction}
\end{figure}
We rewrite the objective defined in Formula \ref{obj-at-K} to be a function of both $K$ which still tracks the index of the current tree we are building and $Z$ which tracks the layer of the tree we are currently building, we now have:
$$L^{(K)} = \sum_i l\left( y_i, \boldsymbol{\hat{y_i}^{K-1)} + f^{Z-1}_{K}(x_i)} + f^{Z}_K(x_i) \right) + \Omega(f^{Z}_K)$$
where $f^{Z}_{K}(x_i)$ is the prediction from the last $Z$ layer of tree $K$ that is currently being built. Notice that one boosting iteration now results in building one layer instead of a whole tree.
Intuitively, the leaves grown in $f^{Z}_K(x_i)$ are learning a residual adjustment over the previous layer. This is useful as deeper trees define more fine grained partitions of the example space and we end up with leaves having few examples where it's desirable to learn smaller adjustments which leverage parent nodes as priors to minimize overfitting.
Furthermore, recalculating gradients at every layer results in better approximating the functional space gradient which in turn typically enables TFBT to build fewer trees due to faster convergence. This is especially true for more complex loss functions, which can be user-defined in TFBT, where the second order Taylor expansion still results in relatively sizable approximation errors.

Additionally, we would like to mention that our implementation allows to choose how many instances to use to recalculate gradients per each layer. On deeper trees, each node would get fewer and fewer examples in a conventional scheme, and splits quality will deteriorate. TFBT allows users to specify the number of instances required for each layer, so the deeper layers can be built on more instances than shallower ones.

\section{TFBT system design}

Below we briefly describe our computation model and changes we had to make to support multi-class learning. For more in-depth review of TFBT please refer to \cite{TfbtEcml}.
\begin{figure*}[ht!]
  \caption{TFBT architecture}
  \centering
  \includegraphics[scale=0.5]{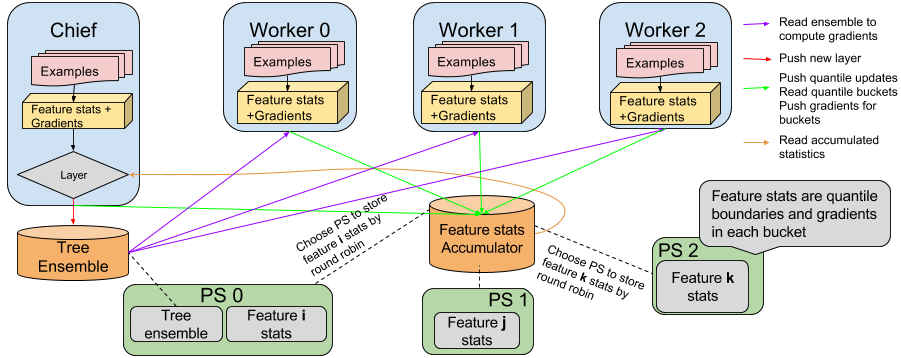}
  \label{figure:architecture}
\end{figure*}

\makeatletter
\def\BState{\State\hskip-\ALG@thistlm}
\makeatother
\algnewcommand\And{\textbf{and}}
\algrenewcommand\alglinenumber[1]{\scriptsize #1:}
\begin{algorithm*}
\scriptsize
\caption{Chief and Workers' work}
\begin{algorithmic}[1]
\Procedure{CalculateStatistics(ps, model, stamp, batch\_data, loss\_fn)}{}
\State $predictions \gets \mbox{model.predict(BATCH\_DATA)}$
\State $quantile\_stats \gets \mbox{calculate\_quantile\_stats(BATCH\_DATA)}$
\State push\_stats(PS,quantile\_stats, stamp) \Comment{\parbox[t]{.35\linewidth}{PS updates quantiles}}
\State $current\_boundaries \gets \mbox{fetch\_latest\_boundaries(PS, stamp)} $
\State $gradients, Hessians \gets \mbox{calculate\_derivatives(predictions,LOSS\_FN)}$
\State $gradients, Hessians \gets \mbox{aggregate(current\_boundaries,gradients, Hessians)}$
\State push\_stats(PS,gradients, Hessians, size(BATCH\_DATA), stamp)
\EndProcedure
\Procedure{DoWork(ps, loss\_fn, is\_chief)}{} \Comment{\parbox[t]{.35\linewidth}{Runs on workers and 1 chief}}
\While {true}
\State $BATCH\_DATA \gets \mbox{read\_data\_batch()}$
\State $model \gets \mbox{fetch\_latest\_model(PS)}$
\State $stamp \gets \mbox{model.stamp\_token}$
\State CalculateStatistics(PS,model, stamp, BATCH\_DATA,LOSS\_FN)
\If{$is\_chief \And \mbox{get\_num\_examples(PS, stamp)} \geq N\_PER\_LAYER} $
\State $next\_stamp \gets stamp+1$
\State $stats \gets \mbox{flush(PS, stamp, next\_stamp)}$ \Comment{\parbox[t]{.35\linewidth}{Update stamp, returns stats}}
\State build\_layer(PS, model,  next\_stamp, stats) \Comment{\parbox[t]{.35\linewidth}{PS updates the ensemble: best splits for the nodes in the layer are chosen according to Formula \ref{eq-split-gain}, spliting is performed}}
\EndIf
\EndWhile
\EndProcedure
\end{algorithmic}
\label{Algo}
\end{algorithm*}
Our design is similar to XGBoost \cite{xgboost} and TencentBoost \cite{tencentboost} in that we build distributed quantile sketches of feature values and use them to build histograms, to be used later to find the best split.
In TencentBoost \cite{tencentboost} and PSMART \cite{psmart} the full training data is partitioned and loaded in workers' memory, which can be a problem for larger datasets. To address this we instead work on mini-batches, updating quantiles in an online fashion without loading all the data into the memory. As far as we know, this approach is not implemented anywhere else.

Each worker loads a mini-batch of data, builds a local quantile sketch, pushes it to a Parameter Server (PS) and fetches the bucket boundaries that were built at the previous iteration. Workers then compute per bucket gradients and Hessians and push them back to the PS. One of the workers, designated as Chief, checks at each iteration if the PS have accumulated enough statistics for the current layer and if so, starts building the new layer by finding best splits for each of the nodes in the layer. Code that finds the best splits for each feature is executed on the PS that have accumulated the gradient statistics for the feature. The Chief receives the best split for every leaf from the PS and grows a new layer on the tree. \\
Once the Chief adds a new layer, the workers copy of the tree ensemble become stale. To avoid stale updates, we introduce an abstraction called StampedResource - a TensorFlow resource with an int64 stamp. The tree ensemble, as well as gradients and quantile accumulators are all stamped resources with such a token. When the worker fetches the model, it gets the stamp token which is then used for all the reads and writes to stamped resources until the end of the iteration. This guarantees that all the updates are consistent and ensures that the Chief doesn't need to wait for Workers for synchronization, which is important when using preemptible VMs (Figure \ref{figure:architecture}). The Chief checkpoints resources to disk and workers don't hold any state, so if they are restarted, they can load a new mini-batch and resume. \\
In order to support multi-class with Diagonal Hessian and Full Hessian strategies, our statistics accumulators support accumulating tensors as well as scalars for each bucket. During TensorFlow graph construction we pick either a tensor accumulator for multiclass or a scalar accumulator for binary classification.
\color{black}

\section{Experiments}
\label{experiments}

In this paper, we make the following claims that we are trying to confirm with experiments.
\begin{compactitem}
\item Our general vector-form multiclass handling is a better way than the conventional 1-vs-rest strategy implemented in most libraries: with a vector-form multiclass, a considerably smaller number of trees is required to reach the best performance. 
\item Layer-by-layer boosting when used with vector form multiclass allows for faster convergences and results in smaller ensembles.
\end{compactitem}
Although not the focus of this paper, we also check (Table \ref{label-mnist-10}) that our implementations are on par with two multiclass vector methods, \textit{LogitBoost} and \textit{ABC-LogitBoost}. We use larger ensembles to be able to fairly compare against publicly available results in \cite{DBLP:journals/corr/abs-1203-3491}.

\subsection{Datasets}
We perform the experiments on medium and large size multiclass datasets, ranging from thousands to a million instances. The details about the datasets and the preprocessing we have done can be found in Table \ref{label:datasets}.
\begin{table*}[ht!]
\centering
\caption{Multiclass datasets descriptions}
\label{my-label}
\renewcommand{\arraystretch}{1}
\setlength{\tabcolsep}{5pt}
\begin{tabular}{clll l p{6cm}}
 \toprule
\textbf{Dataset}  & \textbf{\# train} & \textbf{\# test} &\textbf{\# features} & \textbf{\# classes}  & \textbf{Comments}                                        \\
\midrule
Mnist \cite{mnist}  & 60,000             & 10,000          & 784  & 10  & Hand-written digits recognition task.  Conventional train/test split.  All features are dense numeric. \\

\hline
Sensit \cite{Duarte:2004:VCD:1034812.1034817}  & 78,823 & 19,705 & 100  & 3  &
SensIT Vehicle (combined) dataset, preprocessed data obtained from LibSVM \cite{libsvm} \\

\hline
Covertype \cite{blackard1999comparative}  &  464,810 & 116,202 & 54  & 7  & Obtained from \cite{uci}. Predicting forest cover type. We split all available data into 80\% train and 20\% test.\\

\hline
Letter-26 \cite{frey1991letter}  & 16,000 & 4,000 &16 & 26 & Obtained from \cite{uci}. Conventional test/train split. Predicting a English letter from image features. \\

\hline
Poker \cite{Cattral02evolutionarydata} & 1,000,000 & 25,010 & 10 & 10 & Predicting type of "poker hand" based on information about 5 cards. Preprocessed data obtained from LibSVM \cite{libsvm}  \\

\hline
CIFAR-10 \cite{Krizhevsky09learningmultiple} & 50,000 & 10,000 & 3072 & 10 & Classify RGB 32x32 images into a number of classes. Conventional test/train split. No convolutional features were used.\\

\end{tabular}
\label{label:datasets}
\end{table*}

\subsection{Experiments setup}
We build trees of depth 4 and explore ensembles of 10, 25, 50 and 100 trees. As a baseline we use XGBoost \cite{xgboost} conventional 1-vs-rest multiclass handling. For a default XGBoost model we use the following values of hyperparameters: max\_depth:4, learning\_rate:0.3, objective:multi:softprob, lambda:1, scale\_pos\_weight:False. For tuned XGBoost, we tune min\_child\_weight, learning\_rate and lambda using scikitlearn RandomizedSearchCV \cite{scikit-learn} with 20 iterations with 5-fold CV over the training data using a predefined grid of values. For the poker dataset, 3-fold CV was used due to the fact that there were only 3 instances of class 9. If during hyper parameter tuning, values from the edges of the grid are chosen, the grid is expanded and search is repeated, depth and objective remaining fixed. The grids are as follows:

\begin{lstlisting}[breaklines=true, basicstyle=\footnotesize]
'min_child_weight': [0,0.05,0.1,0.5,1,2,4,8,10,12,14,16,18,20,22],
'learning_rate': [0.01,0.03,0.05,0.075,0.1,0.3,0.5,1,2,3],
'reg_lambda':[0.01, 0.5, 0.1, 1,2,4]
\end{lstlisting}

Note that XGBoost's num\_round parameter (n\_estimators in Scikit-learn) denotes the number of boosting iterations, not the number of trees. During each boosting iteration XGBoost will build $C$ trees. To perform the comparison over the predefined number of trees, we adjust the number of rounds according to following formula:
\begin{lstlisting}[breaklines=true, basicstyle=\footnotesize]
 num_round = int(math.ceil(1.0*num_trees/C))
\end{lstlisting}

In some cases it results in a slightly larger number of XGBoost trees than reported. For example, for \textit{Letter-26} 104 trees are built (for 100 trees column). For the same reasons, it is impossible to build 10 trees for Letter-26 dataset. We don't adjust the number of trees for TFBT implementation (so 10, 25, 50 and 100 trees are built).

For TFBT we use the same objective as XGBoost (namely Max-Ent/Cross entropy/Multinominal Logistic Loss). For TFBT experiments, we use the same hyperparameter values as in default XGBoost: all parameters apart from \textit{lambda} (L2 regularization) translate directly into TFBT settings. We set TFBT's L2 to, since L2 in TFBT is on per instance basis. 
We use full batch size (equal to the training data size), and accumulate train batch size number of instances for layer-by-layer boosting. 

We compare several variations of vector multiclass handling, namely full Hessian, diagonal Hessian and diagonal Hessian with layer-by-layer boosting. We also include the results for conventional tree-per-class (1-vs-rest) implementation, which is also available in TFBT. Note that we don't tune the hyper parameters of our TFBT methods at all, since our goal is to show that vector form multiclass results in smaller ensembles no matter of how much tuning is done.  It should be noted that except in tree-per-class mode, TFBT will use more parameters
per tree than XGBoost, since it stores length $C$ vectors rather than scalars in the leaves.  But we feel that for fixed depth trees, the number of trees is
the more relevant comparison for purposes of inference speed and
understandability.

Finally, to compare with \textit{LogitBoost} and \textit{ABC-LogitBoost}, we replicate Mnist10k (an Mnist dataset with train and test sets reversed) experiment from \cite{DBLP:journals/corr/abs-1203-3491} for depth 2, 3 and 4 and learning rates 0.04, 0.06, 0.08, 0.1. For reasons of space and time, we did not compare against all other
true multiclass approaches mentioned in section \ref{label-related}, but as previously noted, many of them are variations of our formulation.

\subsection{Metrics}
We report the accuracy and the cross entropy loss on each test set. To check for significance between the TFBT and XGBoost results, we use an unpaired t-test
with a p-value threshold of $0.05$.
(We use an unpaired t-test because it was easier to collect the data for, but
since it is less powerful than the paired t-test, if it shows significance,
the paired t-test would have as well.)

\subsection{Experiments results}
\begin{figure}[h!]
  \caption{Accuracy on MNIST}
  \label{fig-mnist}
  \centering
  \includegraphics[scale=0.4]{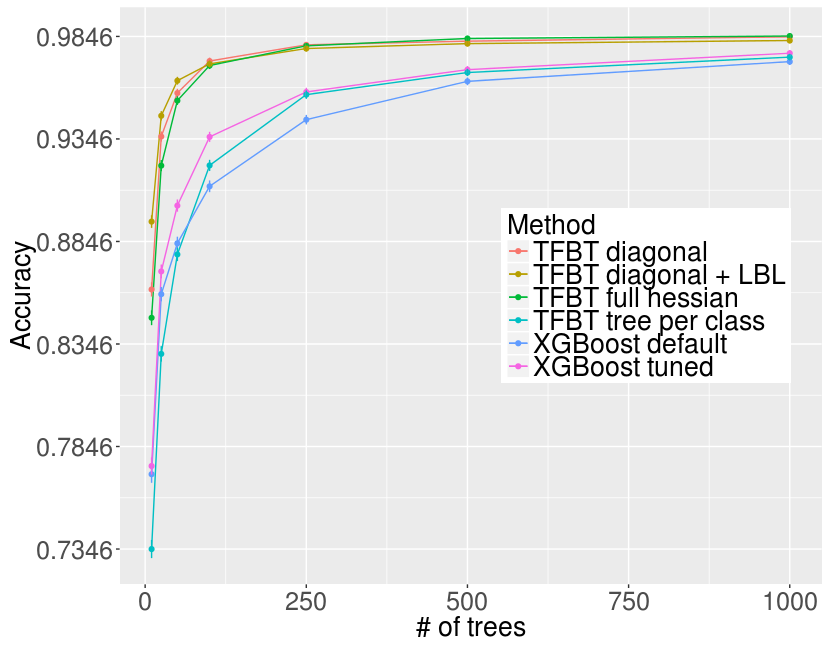}
  \label{figure:architecture}
\end{figure}

\begin{figure}[h!]
  \caption{Accuracy on Letter-26}
  \label{fig-letter}
  \centering
  \includegraphics[scale=0.4]{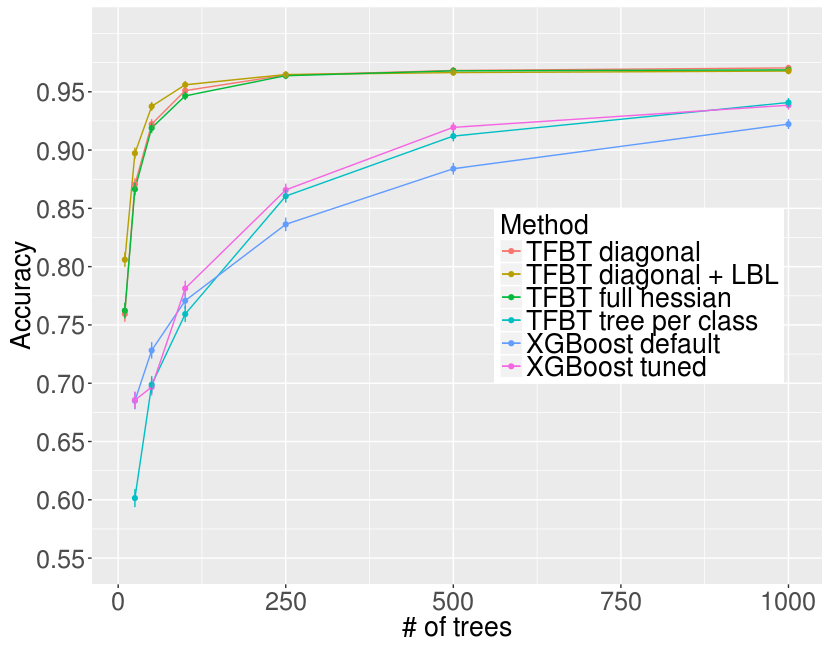}
  \label{figure:architecture}
\end{figure}

Table \ref{multiclass_results} summarizes our results.
Several observations are apparent:
\begin{compactitem}
\item When results are significant, none of the tree-per-class (1-vs-rest) implementations of multiclass handling are able to beat vector multiclass TFBT variants neither in terms of Accuracy nor in terms of Cross-Entropy loss. Sensit dataset is the only dataset where results are non-significant and where tuned XGBoost is able to achieve accuracy and loss that is not significantly different based on the test size and collected standard deviations. One thing to note is that for Sensit, tuned XGBoost ends up using a high learning rate (1 for 10, 25 and 50 trees, and 0.5 for 100 trees), resulting in ``faster'' learning which might lead to overfitting. All TFBT methods still use a learning rate of 0.3.
\item Diagonal Hessian strategy seems to be often significantly better than full Hessian strategy. Since Diagonal version is much faster to run, we recommend to use it by default, instead of the full Hessian method. We also hypothesize that approximating the Hessian matrix via a diagonal matrix serves as a sort of additional regularizer, resulting in better performance in most of the cases.
\item Layer-by-layer boosting (LBL) results in best performance for ensembles of all fixed sizes apart from a single case for Sensit 100 trees. For this particular case, it seems that 100 trees is already enough to achieve best performance even with tree-per-class method, and it suggests that since LBL speeds up convergence, it also may result in overfitting on problems that are easier. 
\item If extremely small ensembles are required (10-50 trees), for example due to inference time or memory size of the device requirements, Diagonal+LBL boosting should be the first choice. Even if larger ensembles can be tolerated, vector form multiclass strategies should be used instead of 1-vs-rest. Figures \ref{fig-mnist} and \ref{fig-letter} present the accuracy convergence on larger number of trees, which shows that even for larger ensembles, vector form multiclass methods dominate. However the line between different vector-form strategies blurs as more trees are added. 
\item The improvements of vector form over tree-per-class, not surprisingly, are more pronounced on datasets with larger number of classes (like Letter-26, Poker, CIFAR-10 and MNIST). It is assumed that approximately $C$ times more trees will be required for tree-per-class to achieve the same performance, and the experiments seem to confirm this folk wisdom.
\item An interesting observation is that TFBT tree-per-class implementation produces different results from XGBoost's version. The difference between implementations lies in the fact that XGBoost computes the predictions for all the instances first and then uses them to construct all $C$ trees in the same boosting iteration, whereas our implementation recalculates the predictions and uses them to calculate the loss and gradients after each subsequent tree. This seems to result in better loss but sometimes leads to worse accuracy (for example, Mnist 10, 25 and 50 trees). Since an improvement in cross-entropy loss does not necessarily translates into a direct improvement in accuracy, we don't find these experiment results alarming. It also demonstrates that recalculating the predictions after each new trees is beneficial to a faster loss convergence, as expected.
\end{compactitem}

\begin{table*}[ht!]
\centering
\caption{Multiclass datasets results}
\label{multiclass_results}
\renewcommand{\arraystretch}{1}
\setlength{\tabcolsep}{5pt}
\begin{tabular}{lllalalala}

\multicolumn{1}{c}{}   & \multicolumn{1}{c}{}   & \multicolumn{2}{c}{\textbf{10 trees}}                                           & \multicolumn{2}{c}{\textbf{25 tres }}   & \multicolumn{2}{c}{\textbf{50 trees }}                                   & \multicolumn{2}{c}{\textbf{100 trees }}      \\

\multicolumn{1}{c}{\multirow{-2}{*}{\textbf{Data}}} & \multicolumn{1}{c}{\multirow{-2}{*}{\textbf{Method}}} & \multicolumn{1}{c}{\textit{Accuracy}}       & \multicolumn{1}{c}{\textit{Cross-Ent}} & \multicolumn{1}{c}{\textit{Accuracy}} & \multicolumn{1}{c}{\textit{Cross-Ent}}  & \multicolumn{1}{c}{\textit{Accuracy}}       & \multicolumn{1}{c}{\textit{Cross-Ent}} & \multicolumn{1}{c}{\textit{Accuracy}}       & \multicolumn{1}{c}{\textit{Cross-Ent}} \\

\hline
\multirow{6}{*}{Mnist} 
& XGBoost default & 0.7711 & 1.5444 & 0.8589 & 1.0151 & 0.8837 & 0.7497 & 0.9115 & 0.4452 \\
& XGBoost tuned & 0.7751 & 0.8479 & 0.8700 & 0.4515 & 0.9021 & 0.3218 & 0.9356 & 0.2084  \\
& Tree-per-class & 0.7346 & 1.2998 & 0.8298 & 0.7830 & 0.8783 & 0.4957 & 0.9217 & 0.2827 \\
& Full Hessian & 0.8474 & 0.5465 & 0.9216 & 0.2613 & \textit{0.9533} & 0.1555 & \textit{0.9704} & \textit{0.0972} \\
& Diagonal Hessian & \textit{0.8612} & \textit{0.4741} & \textit{0.9358} & \textit{0.2201} & \textit{0.9570} & \textit{0.1398} & \textbf{\textit{0.9726}} & \textbf{\textit{0.0917}} \\
& Diagonal + LBL & \textbf{0.8943} & \textbf{0.3513} & \textbf{0.9459} & \textbf{0.1775} & \textbf{0.9630} & \textbf{0.1226} & \textbf{0.9712} & \textbf{0.0899} \\

\hline 
\multirow{6}{*}{Sensit} 
& XGBoost default  & 0.7904                   & 0.6582                   & 0.8091                   & 0.5292                   & 0.8222                   & 0.4691                   & 0.8364                   & 0.4309                   \\
& XGBoost tuned    & \textit{\textbf{0.8088}} & \textit{\textbf{0.4921}} & \textit{\textbf{0.8261}} & \textit{\textbf{0.4456}} & \textit{\textbf{0.8359}} & \textit{\textbf{0.4230}} & \textit{\textbf{0.8430}} & \textit{\textbf{0.4145}} \\
& Tree-per-class   & 0.7836                   & 0.5977                   & 0.8144                   & 0.4869                   & \textit{\textbf{0.8331}} & 0.4399                   & \textit{\textbf{0.8409}} & \textit{\textbf{0.4158}} \\
& Full Hessian     & 0.7998                   & 0.5376                   & \textit{\textbf{0.8245}} & 0.4571                   & \textit{\textbf{0.8386}} & \textit{\textbf{0.4225}} & \textit{\textbf{0.8454}} & \textit{\textbf{0.4060}} \\
& Diagonal Hessian & \textit{\textbf{0.8041}} & \textit{0.4995}          & \textit{\textbf{0.8307}} & \textit{\textbf{0.4407}} & \textit{\textbf{0.8327}} & 0.4431                   & \textit{0.8343}          & 0.4409                   \\
& Diagonal + LBL   & \textbf{0.8110}          & \textbf{0.4833}          & \textbf{0.8285}          & \textbf{0.4442}          & \textbf{0.8391}          & \textbf{0.4235}          & \textbf{0.8407}          & 0.4197     \\
\hline 

\multirow{6}{*}{Covertype}
& XGBoost default  & 0.5991          & 1.3425          & 0.6173                   & 1.1343          & 0.6164          & 0.9669          & 0.6205          & 0.8884          \\
& XGBoost tuned    & 0.6097          & 1.0181          & 0.6158                   & 0.9916          & 0.6072          & 0.9182          & 0.6092          & 0.8982          \\
& Tree-per-class   & 0.7071          & 1.1480          & 0.7264                   & 0.8013          & 0.7362          & 0.6482          & 0.7577          & 0.5668          \\
& Full Hessian     & 0.7113          & 0.6812          & 0.7536                   & 0.5647          & 0.7776          & \textit{0.5085} & \textit{0.8085} & \textit{0.4549} \\
& Diagonal Hessian & \textit{0.7354} & \textit{0.6364} & \textit{\textbf{0.7705}} & \textit{0.5466} & \textit{0.7810} & 0.5361          & 0.7632          & 0.5624          \\
& Diagonal + LBL   & \textbf{0.7399} & \textbf{0.6081} & \textbf{0.7696}          & \textbf{0.5337} & \textbf{0.8047} & \textbf{0.4573} & \textbf{0.8348} & \textbf{0.3927} \\
\hline

\multirow{6}{*}{Letter-26} 
& XGBoost default  &   N/A            &     N/A            & 0.6850          & 1.7549          & 0.7282          & 1.4718          & 0.7708                   & 1.1522          \\
& XGBoost tuned    &      N/A           &     N/A            & 0.6855          & 1.4266          & 0.6967          & 1.1718          & 0.7813                   & 0.8844          \\
& Tree-per-class   & 0.2843          & 2.5872          & 0.6015          & 1.5482          & 0.6988          & 1.1599          & 0.7593                   & 0.8561          \\
& Full Hessian     & \textit{0.7623} & \textit{0.9297} & \textit{0.8665} & \textit{0.5191} & \textit{0.9190} & \textit{0.3118} & \textit{0.9465}          & \textit{0.1879} \\
& Diagonal Hessian & \textit{0.7595} & \textit{0.9263} & \textit{0.8705} & \textit{0.4913} & \textit{0.9223} & \textit{0.2926} & \textit{\textbf{0.9510}} & \textit{0.1800} \\
& Diagonal + LBL   & \textbf{0.8060} & \textbf{0.7339} & \textbf{0.8973} & \textbf{0.3758} & \textbf{0.9375} & \textbf{0.2165} & \textbf{0.9560}          & \textbf{0.1409} \\
\hline

\multirow{6}{*}{Poker} 
& XGBoost default    & 0.5376          & 1.7858                   & 0.5471                   & 1.5421                   & 0.5524          & 1.2046          & 0.5913          & 1.0002          \\
& XGBoost tuned & 0.5376 & 1.1414 & 0.5568 & 0.9871 & 0.5634 & 0.9492 & 0.5880 & 0.9083
 \\
& Tree-per-class     & 0.5086          & 1.7671                   & 0.5212                   & 1.2253                   & 0.5713          & 1.0170          & 0.5966          & 0.9170          \\
& Full Hessian       & \textit{0.5633} & \textit{0.9492}          & 0.6201                   & 0.8567                   & 0.6497          & 0.8089          & 0.6872          & 0.7426          \\
& Diagonal Hessian   & \textit{0.5625} & \textit{\textbf{0.9424}} & \textit{\textbf{0.6301}} & \textit{\textbf{0.8368}} & \textit{0.6606} & \textit{0.7871} & \textit{0.7185} & \textit{0.7012} \\
& Diagonal + LBL     & \textbf{0.5830} & \textbf{0.9299}          & \textbf{0.6351}          & \textbf{0.8249}          & \textbf{0.6750} & \textbf{0.7634} & \textbf{0.7329} & \textbf{0.6651} \\
\hline

\multirow{6}{*}{CIFAR-10} 
& XGBoost default  & 0.3012                   & 2.1725          & 0.3528                   & 2.0195          & 0.3769                   & 1.9221                   & 0.4050                   & 1.7724                   \\
& XGBoost tuned    & 0.3003                   & 2.0021          & 0.3546                   & 1.9215          & 0.3829                   & 1.7361                   & 0.4194                   & 1.6361                   \\
& Tree-per-class   & 0.2885                   & 2.1113          & 0.3285                   & 1.9709          & 0.3622                   & 1.8335                   & 0.4053                   & 1.6967                   \\
& Full Hessian     & \textit{0.3450}          & \textit{1.8574} & \textit{\textbf{0.4085}} & \textit{1.6830} & \textit{\textbf{0.4507}} & \textit{\textbf{1.5573}} & \textit{\textbf{0.4856}} & \textit{\textbf{1.4601}} \\
& Diagonal Hessian & \textit{\textbf{0.3568}} & \textit{1.8418} & \textit{\textbf{0.4068}} & \textit{1.6746} & \textit{\textbf{0.4497}} & \textit{\textbf{1.5674}} & \textit{\textbf{0.4798}} & \textit{\textbf{1.4778}} \\
& Diagonal + LBL   & \textbf{0.3631}          & \textbf{1.7933} & \textbf{0.4148}          & \textbf{1.6458} & \textbf{0.4548}          & \textbf{1.5434}          & \textbf{0.4769}          & \textbf{1.4903}         \\
\hline
 \multicolumn{10}{l}{\footnotesize 
\parbox[t]{.9\linewidth}{
 For each dataset and column, in \textbf{bold} are the best values of the metrics for cross entropy-loss and accuracy for this number of trees. In \textit{italics}, we highlight the best value of cross-entropy and accuracy when Diagonal+LBL is not considered. If there is no significant difference between several best values, we highlight all the values the difference between which is insignificant.}} \\

\end{tabular}
\end{table*}

\begin{table}[ht!]
\centering
\caption{Comparison of accuracy on Mnist10k}
\label{label-mnist-10}
\renewcommand{\arraystretch}{1}
\begin{tabular}{llccccccc}
\textbf{LR}           & \textbf{D} & \textbf{LogitBoost} & \textbf{ABC-Logit} & \textbf{Diagonal} & \textbf{Diag+LBL}  \\
\hline
\multirow{3}{*}{0.04} & 2              & 0.9511              & 0.9562                  & 0.9562                 & 0.9585            \\
                      & 3              & 0.9567              & 0.9640                  & 0.9643                 & 0.9617                       \\
                      & 4              & 0.9596              & 0.9648                  & 0.9661                 & 0.9646	 \\
                      
\hline
\multirow{3}{*}{0.06} & 2              & 0.9505              & 0.9567                  & 0.9768                 &                            0.9556 \\
                      & 3              & 0.9564              & 0.9644                  & 0.9640                 & 0.9614                      \\
                      & 4              & 0.9594              & 0.9648                  & 0.9657                 &                            0.9637 \\
                      
\hline
\multirow{3}{*}{0.08} & 2              & 0.9503              & 0.9578                  & 0.9582                 &                            0.9561 \\
                      & 3              & 0.9569              & 0.9647                  & 0.9642                 & 0.9615                      \\
                      & 4              & 0.9601              & 0.9651                  & 0.9660                 & 0.9617                      \\
                      
\hline                     
\multirow{3}{*}{0.1}  & 2              & 0.9497              & 0.9580                  & 0.9588                 &          0.9561       \\
                      & 3              & 0.9567              & 0.9643                  & 0.9640                 &                             0.9614 \\
                      & 4              & 0.9601              & 0.9653                  & 0.9660                 &                           0.9635 \\

\hline
\multicolumn{6}{l}{\footnotesize 
\parbox[t]{.9\linewidth}{\textit{D} stands for depth, \textit{LR} for learning rate, \textit{ABC-Logit} is ABC-LogitBoost, \textit{Diagonal} and \textit{Diag+LBL} are TFBT implementations. Results for LogitBoost and ABC-LogitBoost are from \cite{DBLP:journals/corr/abs-1203-3491} Table 2.} }
 
\end{tabular} 
\end{table}

\section{Conclusion}
We show that the conventional boosting formalism used by most popular open-sourced gradient boosting libraries can be easily extended from scalar-valued trees to vector-valued trees. We demonstrate, not surprisingly, that vector-valued trees lead to much faster convergence and smaller ensembles. We also show that other recent vector-valued gradient boosted trees formulations fit into our general framework and can be easily implemented in TFBT by changing the loss, Hessian handling strategy and label encoding scheme. We would like to encourage researchers who work on multiclass boosting to reuse our TensorFlow based TFBT library, which is open-sourced and convenient to use due to automatic differentiation capabilities. Finally, we argue that vector-valued leaves should be the default strategy for handling multiclass problems.

\section*{Acknowledgment}
The authors would like to thank Boris Dadachev, Afshin Rostamizadeh, Corinna Cortes, Tal Shaked, D. Sculley, Alexander Grushetsky and Petr Mitrichev for their invaluable comments on this paper.



%

\bibliographystyle{IEEEtran}
\bibliography{IEEEabrv,arxiv_version}

\end{document}